\documentclass[conference]{IEEEtran}
\IEEEoverridecommandlockouts
\usepackage{cite}
\usepackage{amsmath,amssymb,amsfonts}
\usepackage{algorithmic}
\usepackage{graphicx}
\usepackage{multirow}
\usepackage{siunitx}
\usepackage{textcomp}
\usepackage{xcolor}
\usepackage{subfigure}
\def\BibTeX{{\rm B\kern-.05em{\sc i\kern-.025em b}\kern-.08em
    T\kern-.1667em\lower.7ex\hbox{E}\kern-.125emX}}

\providecommand{\DIFdel}[1]{} % Don't show deleted text

\begin{document}

% 校正者の評価について

% 最終確認について

\title{Roadside-assisted Cooperative Planning using Future Path Sharing for Autonomous Driving
% {\footnotesize \textsuperscript{*}Note: Sub-titles are not captured in Xplore and
% should not be used}
% \thanks{Identify applicable funding agency here. If no, delete this.}
}

\author{\IEEEauthorblockN{
    Mai Hirata\IEEEauthorrefmark{1}, 
    Manabu Tsukada\IEEEauthorrefmark{1},
    Keisuke Okumura\IEEEauthorrefmark{2},
    Yasumasa Tamura\IEEEauthorrefmark{2},
    Hideya Ochiai\IEEEauthorrefmark{1},
    Xavier D\'efago\IEEEauthorrefmark{2}
}
\IEEEauthorblockA{\IEEEauthorrefmark{1}
Graduate School of Information Science and Technology, The University of Tokyo, Tokyo, Japan\\
Email: \textit{\{mai, tsukada, jo2lxq\}@hongo.wide.ad.jp}}
\IEEEauthorrefmark{2}
School of Computing, Tokyo Institute of Technology, Tokyo, Japan\\
Email: \textit{\{okumura.k, tamura.y, defago.x\}@coord.c.titech.ac.jp}
}

\maketitle

\begin{abstract}
    Cooperative intelligent transportation systems (ITS) are used by autonomous vehicles to communicate with surrounding autonomous vehicles and roadside units (RSU).
    Current C-ITS applications focus primarily on real-time information sharing, such as cooperative perception.
    In addition to real-time information sharing, self-driving cars need to coordinate their action plans to achieve higher safety and efficiency.
    For this reason, this study defines a vehicle’s future action plan/path and designs a cooperative path-planning model at intersections using future path sharing based on the future path information of multiple vehicles.
    The notion is that when the RSU detects a potential conflict of vehicle paths or an acceleration opportunity according to the shared future paths, it will generate a coordinated path update that adjusts the speeds of the vehicles. 
    We implemented the proposed method using the open-source Autoware autonomous driving software and evaluated it with the LGSVL autonomous vehicle simulator.
    We conducted simulation experiments with two vehicles at a blind intersection scenario, finding that each car can travel safely and more efficiently by planning a path that reflects the action plans of all vehicles involved. The time consumed by introducing the RSU is \SI[detect-weight]{23.0}{\percent} and \SI[detect-weight]{28.1}{\percent} shorter than that of the stand-alone autonomous driving case at the intersection.    
\end{abstract}

% PageReduction
% \begin{IEEEkeywords}
% Autonomous vehicles, Cooperative intelligent transport systems, Cooperative planning, Open-source software, Vehicle-to-everything 
% \end{IEEEkeywords}

\section{Introduction}
% ITS
Autonomous driving technologies are a part of intelligent transport systems (ITS) that use information and communication technologies to exchange pedestrian, road, and vehicles data to mitigate common road and traffic problems, including accident and congestion avoidance. Consequently, they have attracted a great deal of attention in recent years.
Car companies have already commercialized related functions, such as autonomous emergency braking, lane-keeping assistance, and automatic parking~\cite{HondaGlo30:online,ToyotaBr10:online}. Several organizations are developing autonomous automated driving along with open-source software packages such as Autoware \cite{7368032} and Apollo~\cite{apollo:github}. 

% According to the Japanese Ministry of Land, Infrastructure, Transport and Tourism (MLIT), 66.2 \% of new vehicles use collision damage mitigation braking as of 2016. 
% Society of Automotive Engineers (SAE) defines these features into Level 1 or 2.
% In Japan, Honda received the type designation for Level 3 by MILT in November 2020~\cite{HondaGlo30:online}.

% Cooperative ITS

Cooperative ITS is important because autonomous vehicles cannot otherwise guarantee the safety of pedestrians, vehicles, and property. 
The common architecture is a station infrastructure~\cite{ISO-21217-CALM-Arch,ETSI-EN-302-665-Arch} that was developed by the European Telecommunications Standards Institute (ETSI) and the International Organization for Standardization (ISO). According to standards, a cooperative awareness message (CAM)~\cite{ETSI-EN-302-637-2-CAM} is the fundamental safety-related vehicle-to-vehicle (V2V) process that shares real-time data about other vehicles. 
% In Toyota's ITS Connect~\cite{ToyotaBr10:online}, ITS dedicated wireless devices are installed at some of the intersections with heavy traffic throughout Japan. 
% ITS Connect provides right-turn alerts and red-light alerts at these intersections.
%
% Cooperative ITS also needs to address legacy vehicles that are not equipped with Vehicle-to-everything (V2X) transmitters and vulnerable road users such as pedestrians and bicycles. 
Vehicle-to-everything (V2X) processes are advancements that share sensors information to vehicles and roadside equipment to create a collective perception capability.
In this regard, ETSI is in the process of standardizing collective perception messages (CPMs)~\cite{ETSI-TS-103-324-CPS,ETSI-TR-103-562-CPS}.

% issues in our research 
The future information consists of a list of planned positions of a vehicle over time (i.e., the planned trajectory). According to ETSI, a maneuver coordination message (MCM) exchanges planned trajectories and performs driving coordination. However, its format is still under development~\cite{ETSI-TR-103-578-MCM}, and no standard has been issued. Maneuver coordination (MC) research has been conducted in various scenarios such as platooning, lane changing, and merge coordination\cite{Lehmann2018-bb, Wang2018-ef}. 

In many field demonstrations, infrastructure-based CPM has been demonstrated to improve the perception of automated vehicles\cite{Shan2020-co, Tsukada2020}. 
However, research on the infrastructure which supports the vehicle's better prediction and planning is still in its nascent stage. Especially it is not thoroughly investigated with the full-fledged autonomous software. 

The contribution of this study is designing a roadside unit (RSU) that enhances vehicle' prediction and planing cooperatively with other vehicles to improve traffic flow at intersections. We implemented the solution in Autoware~\cite{7368032}, open-source software for autonomous driving systems. 
To the best of our knowledge, this study is the first to implement cooperative planning in Autoware. 
The proposed method is divided into two schemes: a periodic message broadcast to exchange vehicles' future path (\emph{i.e., future path sharing}) and a maneuver coordination algorithm based on a reservation table that manages the set of future paths in the RSU (\emph{i.e., cooperative planning}). 

% We define a mechanism for exchanging future route information. 
% We propose a method for bundling this information and managing future routes by RSU at the intersection.
% We implement the proposed method in Autoware~\cite{7368032}, open-source software for autonomous driving systems, and evaluated it in the LGSVL simulator~\cite{LGSVLSim89:online}.

% Paper organization
The rest of this paper is structured as follows.
% PageReduction
% Section~\ref{sec:related-work} reviews related works related to cooperative perception, cooperative planning, maneuver coordination, and intersection management. 
% Section~\ref{sec:proposition} describes our proposal for future path sharing-based cooperative planning. 
% Section~\ref{sec:implementation} shows the details of our implementation based on Autoware. 
% Section~\ref{sec:evalation} evaluates the implementation using LG Corporation’s Silicon Valley Lab (LGSVL) autonomous vehicle simulator ~\cite{Rong2020-ie}.
% Section~\ref{sec:conclusion} concludes the paper and provides future works. 
Section~\ref{sec:related-work} reviews related works and 
Section~\ref{sec:proposition} describes our proposal. 
Section~\ref{sec:implementation} shows our implementation based on Autoware. 
Section~\ref{sec:evalation} evaluates the implementation using LG Corporation’s Silicon Valley Lab (LGSVL) simulator ~\cite{Rong2020-ie}.
Section~\ref{sec:conclusion} concludes the paper and provides future works.

\section{Related work}\label{sec:related-work}

Renzler et al.~\cite{PathFuture_Renzler2020} proposed CAM extensions to inform future path information for emergency response in the event of collisions. It was shown to reduce evasive maneuvers that would otherwise pose a danger to vehicles in adjacent lanes.

Technologies for sharing sensor information, including ITS messages, can be broadly divided into three levels: low, feature, and track ~\cite{aeberhard2011high}.
% At the low level, raw data is shared.
% At the feature level, specific features are extracted from the raw data as a pre-processing step before sharing the data.
% At the track level, each sensor independently runs a tracking algorithm to generate an obstacle list. Then, the obstacle list is shared.
%
Track-level data exchange has the advantage of not requiring large network bandwidth.
Gabb et al.~\cite{Gabb2019-gw} presented a scheme for data sent from a mobile edge computing (MEC) server to a car and proposed a system that fuses data from the car's sensors with an environmental model based on infrastructure sensor data at the MEC server.
Tsukada et al.~\cite{Tsukada2020,Tsukada2020b} proposed AutoC2X, a system that shares information collected by roadside equipment with an automatic vehicle using Autoware~\cite{7368032} and OpenC2X, open-source experimental and prototype platforms for cooperative ITS.

Existing research on the management of autonomous vehicles at intersections can be categorized into decentralized and centralized approaches.
In the decentralized approach, autonomous vehicles communicate via a wireless network.
Azimi et al.~\cite{6843706} proposed a cooperative intersection management algorithm that models the area of an intersection as a grid.
Each cell in the grid is assigned a unique identifier and shares current and future locations with vehicles.
Aoki et al.~\cite{8443745} studied how autonomous vehicles drive through dynamic intersections in ways that lead to accidents using sensor-based perception and inter-vehicle communication.
The autonomous vehicles communicate with the Intersection Manager (IM) for the crossing plan for the centralized approach.
Dresner et al.~\cite{dresner2008multiagent} proposed autonomous intersection management (AIM) system for coordinating the independent movements of self-driving cars through intersections.
Liu et al.~\cite{LIU201916} proposed a trajectory-planning-based version, whereas Bashiri et al.~\cite{7995794} proposed a platoon-based approach.

% mannouver coordination
Maneuver coordination (MC) research can be divided into two categories: scenario-specific MC and general MC. 
In a platooning coordination study, \cite{englund2016grand} suggested a V2X protocol for handling all possible scenarios of platooning. Vehicles are coordinated with the protocol to merge two platooning sequences into one. \cite{Sawade2018-rj} also proposed a redundant message protocol for cooperative driving that involves various vehicles, including platooning. In the proposed system \cite{Milanes2011-bl} for smooth ramp merging, the roadside unit at the merging point detects the merging activity while searching for vehicles on the main road with the possibility of involving in a collision. Subsequently, it instructs speed adjustment to the identified vehicles. Furthermore, the cooperative maneuver protocol (CMP) is proposed \cite{Eiermann2020-np} as a coordination message at the merge point. For coordinated lane changing, \cite{Wang2018-ef} suggested lane-change messaging. Various studies have employed the space-time reservation protocol (STRP) that uses messaging for reserving a particular location for a certain period. \cite{Heb2018-dx} used STRP for lane-change coordination and carried out field experiments to justify the technique. In \cite{Nichting2020-zy}, the method is extended to intersections, passing, and roundabouts.

\cite{Schindler2019-lc} put forward the TransAID project that conducts research and development on more generic V2X messages. It also involved the study of cooperative automated driving and Cooperative ITS standard messages, including CAM, CPM, and MCM~\cite{correa2019infrastructure, correa2019transaid}. \cite{correa2019infrastructure} suggested a format for a more generic MCM message. It considered the transition of control (switching from automatic to manual driving) as an instruction type for MCM along with lane change and speed adjustment. In \cite{correa2019transaid}, a V2X framework was implemented for exchanging each message sets standardized in Cooperative ITS.

Furthermore, in~\cite{Jacob_undated-ga, auerswald2019cooperative}, a full stack from the access layer was applied to the facility layer. Besides MCM, they used a maneuver recommended message for driving coordination. For the same purpose, \cite{Hafner2020-xn} proposed the complex vehicular interactions protocol. \cite{Decentra11:online} assumed that vehicles always provide a trajectory when carrying out driving cooperation. Apart from using MCM for the reduction of uncertainty in automated vehicles, \cite{Lehmann2018-bb} proposed the MCM as a more general protocol. 
An Autoware-based MC protocol is implemented as AutoMCM~\cite{Mizutani2021b}, and the system demonstrates robustness against packet loss in the experiments.

We reviewed these studies and others on future path replacement, driving management at intersections, and maneuver coordination.
Future path exchange methods are in their infancy, and no experimental data have yet been obtained.
There have also been just a few examples of intersection driving management platforms that enable actual autonomous driving.
% PageReduction
% Therefore, in this research, we propose a method of intersection driving management based on future path exchange. We implement it for real autonomous driving and conduct experiments using a simulator.

% \section{Problem Statement}
% We describe the issues of cooperative planning.

\section{Future Path Sharing based Cooperative Planning}\label{sec:proposition}

\begin{figure}[tbp]
    \begin{center}
        \includegraphics[width=0.7\linewidth,clip]{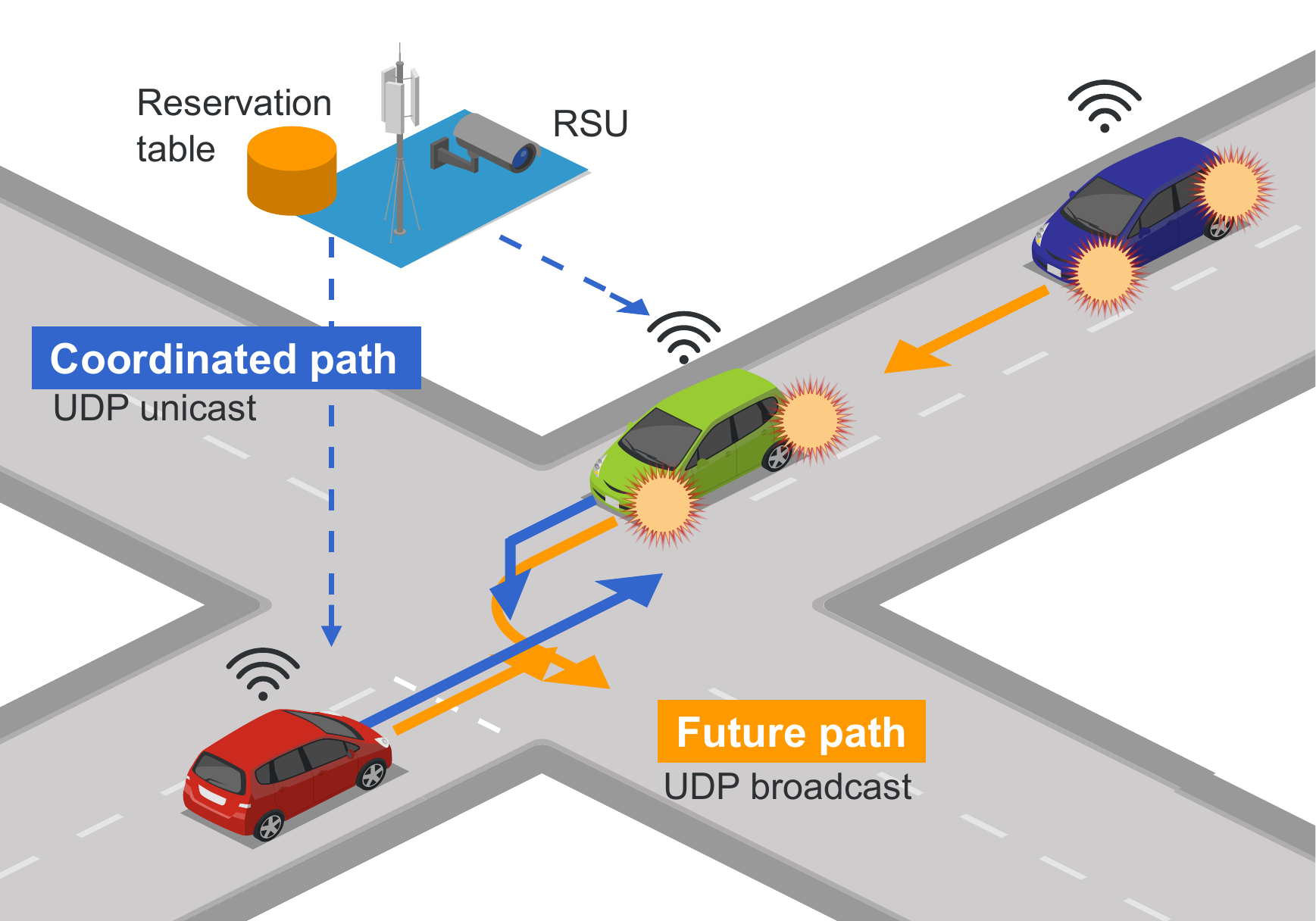}
        \caption{Cooperative planning using future path sharing}
        \label{fig:proposed_model}
    \end{center}
\end{figure}

Figure~\ref{fig:proposed_model} shows an overview of our scheme. 
The autonomous vehicles periodically broadcast their \textit{future path} information to the surrounding vehicles. 
Receiver vehicles make adjustments based on the perception of the other vehicles' future paths as obstacles. 
As a result, all vehicles that receive the future path will slow down and not pass through the intersection efficiently.
In our proposal, the RSU improves the situation by collecting future paths and adding them to its \textit{reservation table}. 
When one future path conflicts with another vehicles' future path in the reservation table, the RSU computes a conflict-free resolution (\textit{coordinated path}) and sends it to the affected vehicle. 
When the vehicle receives a new path from the RSU, it prioritizes it over its previously prepared path. 
In the following sections, we describe future path sharing and coordinated path generation in detail.

\subsection{Future path sharing}

\begin{figure}[tbp]
    \begin{center}
        \includegraphics[width=0.72\linewidth,clip]{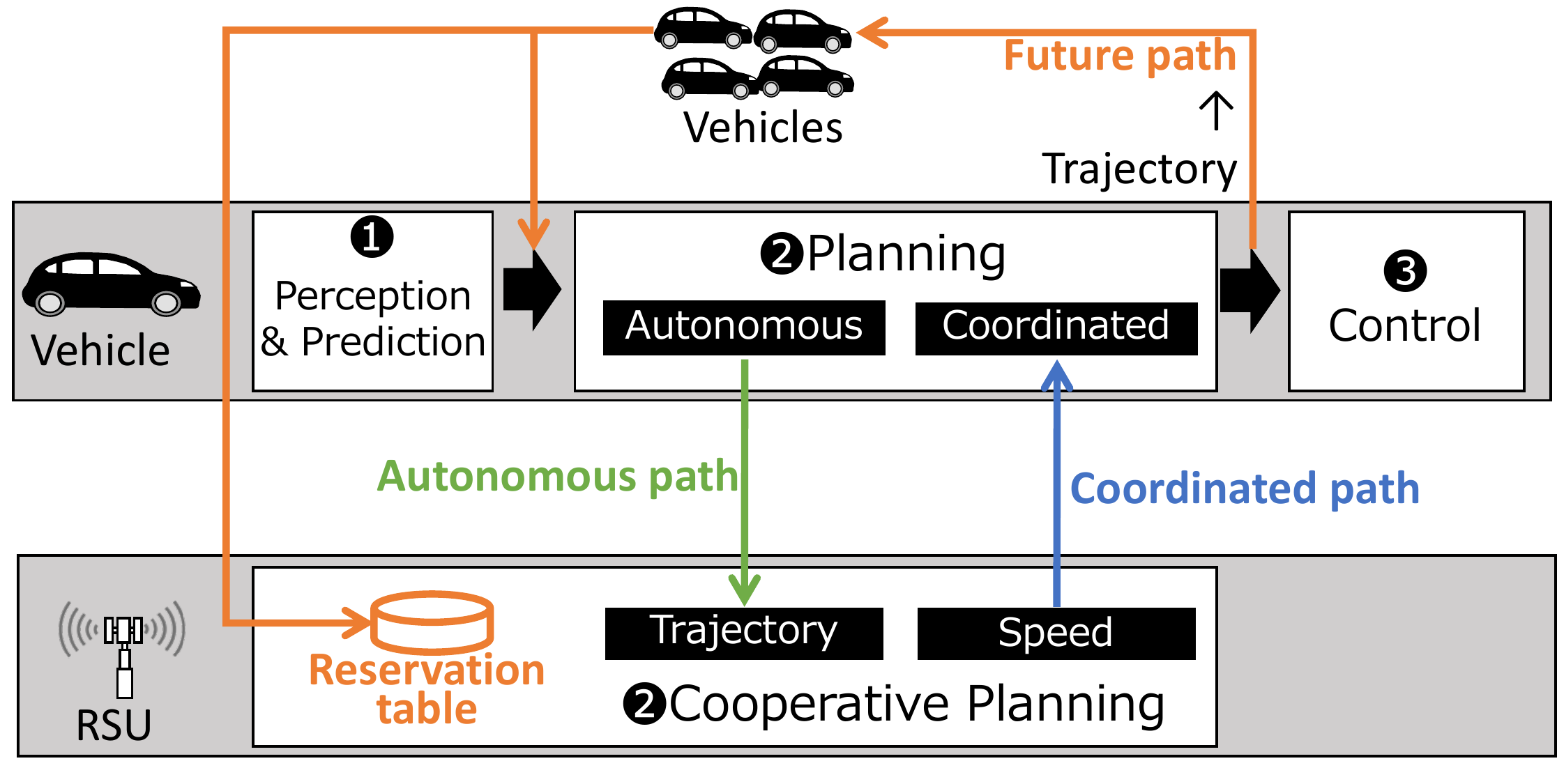}
        \caption{Future path, autonomous path, coordinated path, and reservation table}
        \label{fig:3_paths}
    \end{center}
\end{figure}

Figure~\ref{fig:3_paths} displays a typical autonomous driving system. 
First, the perception subsystem processes sensor inputs to interpret the surrounding environment. 
It detects the location, size, and type of surrounding vehicles and pedestrians and predicts their movements.
The planning module then calculates a route from the map to the target point and generates an action plan along that route.
As an output, the planning module generates a trajectory and provides it to the control module.

We define the following terms to describe our system:
\begin{itemize}
    \item \textbf{The trajectory} consists of continuous paired sequences of the location and velocity. 
    It results from the planning module's decision, and the control module follows the trajectory in typical autonomous driving software (e.g., Autoware). 
    \item \textbf{The future path} consists of continuous paired sequences of coordinates and timing factors of their passing. The future path employs a data structure that makes it easy for cars and RSUs to detect conflicts.
    We generate the future path assuming that the velocity from a position to the next position is constant. 
The time to pass $n-$th point is calculated by equation~\eqref{eq}.
\begin{equation}
t_n=\sum_{k=0}^{n} \frac{x_k-x_{k-1}}{v_k} + t_0 (n\geq1),
\label{eq}
\end{equation}
where ${t_0}$ is the current time, $x_0$ and $v_0$ are the pair of the location and the velocity of the closest point to the current vehicle position in the trajectory. 
\end{itemize}

Our proposal converts the trajectory output from the planning module to the future path using equation~\eqref{eq}.
Then, the future path is broadcast to the surrounding vehicles with an identifier, the current position, the current speed, and the vehicle shape.
The receiving vehicle can treat the received information as a dynamic obstacle by integrating it with the output of the perception module. 
This integration potentially causes unnecessary deceleration because the future path is treated as a future obstacle. 
Therefore, a coordinated path (detailed in the next section) is necessary to improve traffic flow at the intersection. 

% In addition to the local sensor's perception and prediction on the vehicle, we propose to exchange \emph{future paths} between vehicles via wireless communication. 

% Each autonomous vehicle calculates the time to pass through each point and generates a \emph{future path}, as equation~\eqref{eq}.

\subsection{Coordinated path generation}

We define the following terms to describe our system:

\begin{itemize}
    \item \textbf{The reservation table} is the RSU's database that stores the set of future paths sent by autonomous vehicles. The reservation table is used to detect conflicts between cars and calculate opportunities for acceleration.
    \item \textbf{An autonomous path} is calculated by the autonomous vehicle to the desired destination. 
    The vehicle periodically unicasts its autonomous path to the RSU during the coordinated mode. 
    Expectedly, the path may conflict with other vehicles paths, causing unnecessary deceleration.
    \item \textbf{The coordinated path} is the new path in which the RSU modifies the autonomous path to avoid conflicts and improve traffic flow. The RSU periodically unicasts this information to the target vehicles, and the vehicles give priority to this path.
\end{itemize}

Our scheme is a hybrid one in which vehicles create a trajectory, and the RSU modifies their speeds along the trajectories. 
The RSU receives the vehicles' future paths, as shown in Fig.~\ref{fig:3_paths}, and it updates its reservation table. 
When a received future path conflicts with another path in the table, the RSU sends an initiation message to a vehicle to enter a coordinated mode. 
As depicted in Fig.~\ref{fig:3_paths}, the RSU then generates a coordinated path that slows the vehicle's speed until the future path is sufficiently shortened, and the conflict disappears. This could involve one vehicle ultimately stopping.  
% In other words, the vehicle decides the trajectory, and RSU determines the speed in our scheme. 
% Simultaneously, the vehicle in the coordinated mode sends the \emph{autonomous path} to the RSU because the vehicle's \emph{future path} is identical to the \emph{coordinated path} in this mode. 
% And \emph{autonomous path} is necessary to take the desired destination of the vehicle. 

When the received future path does not conflict with any path in the reservation table, the RSU determines room for acceleration. If so, it performs the same steps, resulting in the vehicle temporarily speeding up.

% RSUs are responsible for managing the \emph{future paths} sent by autonomous vehicles.
% The management method is to have a bundle of \emph{future paths}.

% We assumed that the number of intruding vehicles at an intersection is limited, and we managed the \emph{future paths} as a bundle.
% By managing \emph{future paths} in bundles, we can determine which vehicle the managed \emph{future path} belongs to, making it easier to determine if the \emph{future paths} conflict.

% When the vehicle runs according to the coordinated trajectory, it sends the autonomous future path to the RSU in unicast at a cycle of 10Hz.

% When the car passes through the intersection, the RSU allows the car to drive according to the autonomous trajectory. At this point, the autonomous future path will again be broadcast as the selected future path.

% % \subsection{Integration of future path and predictions}
% Pedestrians, who cannot send a \emph{future path}, are managed using trajectory predictions.
% The RSU predicts the trajectory of the recognized pedestrian and generates a \emph{future path} based on the prediction.
% Then, the RSU adds the \emph{future path} to the reservation table.

\section{Implementation}\label{sec:implementation}

We implemented the proposed method by extending Autoware.IV (v.0.6.0). In this section, we describe Autoware and explain our proposed extension in detail. 

\subsection{Autoware}
Autoware is a Linux-based open-source software for autonomous driving~\cite{7368032} and is based on the Robot Operating System (ROS)~\cite{quigley2009ros}.
% ROS
ROS is a distributed robot platform that utilizes \textit{nodes} and \textit{topics}. The nodes represent the processing module of a task and interact via topics. The ROS includes the \textit{RViz} 3D visualization tool, which displays the status of the tasks.
Autoware provides a set of applications necessary for autonomous driving, including localization, perception, planning, and control. 
Using 3D maps, LiDAR, a camera, and the Global Navigation Satellite System, Autoware performs localization and perception. 

% https://github.com/tier4/AutowareArchitectureProposal.proj/blob/master/design/Planning/Planning.md
% https://github.com/tier4/AutowareArchitectureProposal.proj/blob/master/design/Planning/DesignRationale.md
% https://github.com/tier4/AutowareArchitectureProposal.proj/blob/master/design/Planning/ScenarioSelector/ScenarioSelector.md
% https://github.com/tier4/AutowareArchitectureProposal.proj/blob/master/design/Planning/LaneDriving/LaneDrivingScenario.md
% https://github.com/tier4/AutowareArchitectureProposal.proj/blob/master/design/Planning/Parking/ParkingScenario.md
The perception module provides detected obstacle information to the Autoware planning module, which is divided into three parts: mission planning, scenario selection, and behavior planning.
The mission planner calculates the entire route to the goal by searching a static map. 
The scenario selector decides which behavior planner should be applied depending on the situation. Notably, it is technically challenging to have a unified behavior planner handle every possible situation. Currently, Autoware provides two: one for on-road driving and another for parking. 
The selected planner gives the calculated trajectory to the control module that manipulates acceleration, braking, and steering via a controller area network. 

% Lanelet2~\cite{Poggenhans2018-fk}

\subsection{Implementation overview}

\begin{figure}[tbp]
    \begin{center}
        \includegraphics[width=\linewidth,clip]{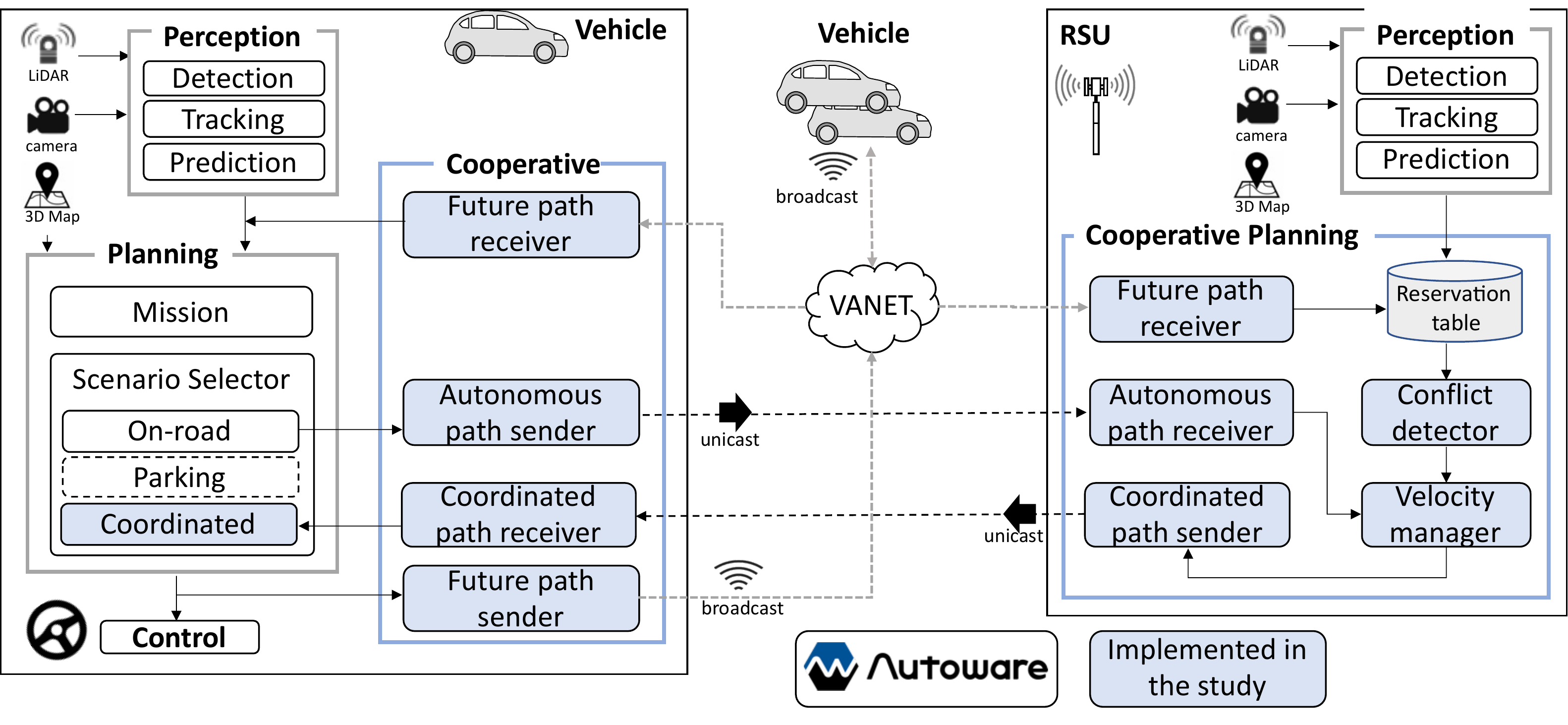}
        \caption{Implementation overview}
        \label{fig:vehicle_planning}
    \end{center}
\end{figure}

Figure~\ref{fig:vehicle_planning} shows the overview of our implementation. The white box comes with Autoware, and the blue box is newly implemented in our extension. 

We modified the structure of ROS nodes as follows. We added the coordinated planner in addition to the on-road planner and the parking planner.
The scenario selector switches between the on-road and coordinated planners depending on the mode (described later). The parking planner is not used. The planning module passes the trajectory to the control module with the topic, \mbox{\textsf{\//planning\//scenario\_planning\//trajectory}}. 
The \textbf{future-path sender} converts the trajectory to a future path and sends it using a user datagram protocol (UDP) broadcast at a frequency of \SI{10}{\hertz}. 
A future path includes 120 trajectory points, and the distance between each point should be at least \SI{0.1}{\meter}. The network load and the implications are analyzed in section~\ref{subsec:discussion}. 
Both vehicles and the RSU receive the future path, the future path receiver publishes it to \mbox{\textsf{\//perception\//object\_recognition\//objects}}, and the perception module sends the surrounding objects to the planning module. Hence, the planning module can also employ future data from V2X. 

When the RSU receives the data and no conflicts are detected, the future path is added to the reservation table. 
In case of a conflict, the RSU sends an initiation message to the vehicle requesting it switch to the coordinated mode. 
The vehicle then activates the \textbf{autonomous path sender} to send its autonomous paths to the RSU periodically at a unicast of \SI{100}{\milli\second} intervals. 
Upon reception at the RSU, the velocity manager generates a coordinated path by modifying the autonomous path's speed from the autonomous path receiver. The velocity manager then decreases the vehicle's speed when a conflict is detected and increases its speed when it discovers room for acceleration. When the vehicle receives the coordinated path from the RSU, the vehicle performs planning in the coordinated mode. 

% 'PageReduction'
% The RSU can take advantage of the Autoware perception module to detect all objects near the intersection, including non-connected cars and vulnerable road users (e.g., cyclists and pedestrians). 
% The RSU then adds the future path shared in \mbox{\textsf{\//perception\//object\_recognition\//objects}} to the reservation table. From the newly added paths, the RSU generates coordinated paths to avoid conflicts. 

As a limitation of the current implementation, it only focuses on the longitudinal vehicle motion, and the lateral vehicle motion calculation is future work.
Also, our current implementation uses the JavaScript Object Notation (JSON) format~\cite{rfc7159} to send the future path, the autonomous path, and the coordinated path where the size of the data is 60 KB when the path has 120 points. 
The future work includes reducing the packet size within the maximum transmission unit size of UDP (i.e., 1460 Bytes).

\subsection{Coordinated path generation with mode management}

\begin{figure}[tbp]
    \begin{center}
        \includegraphics[width=\linewidth,clip]{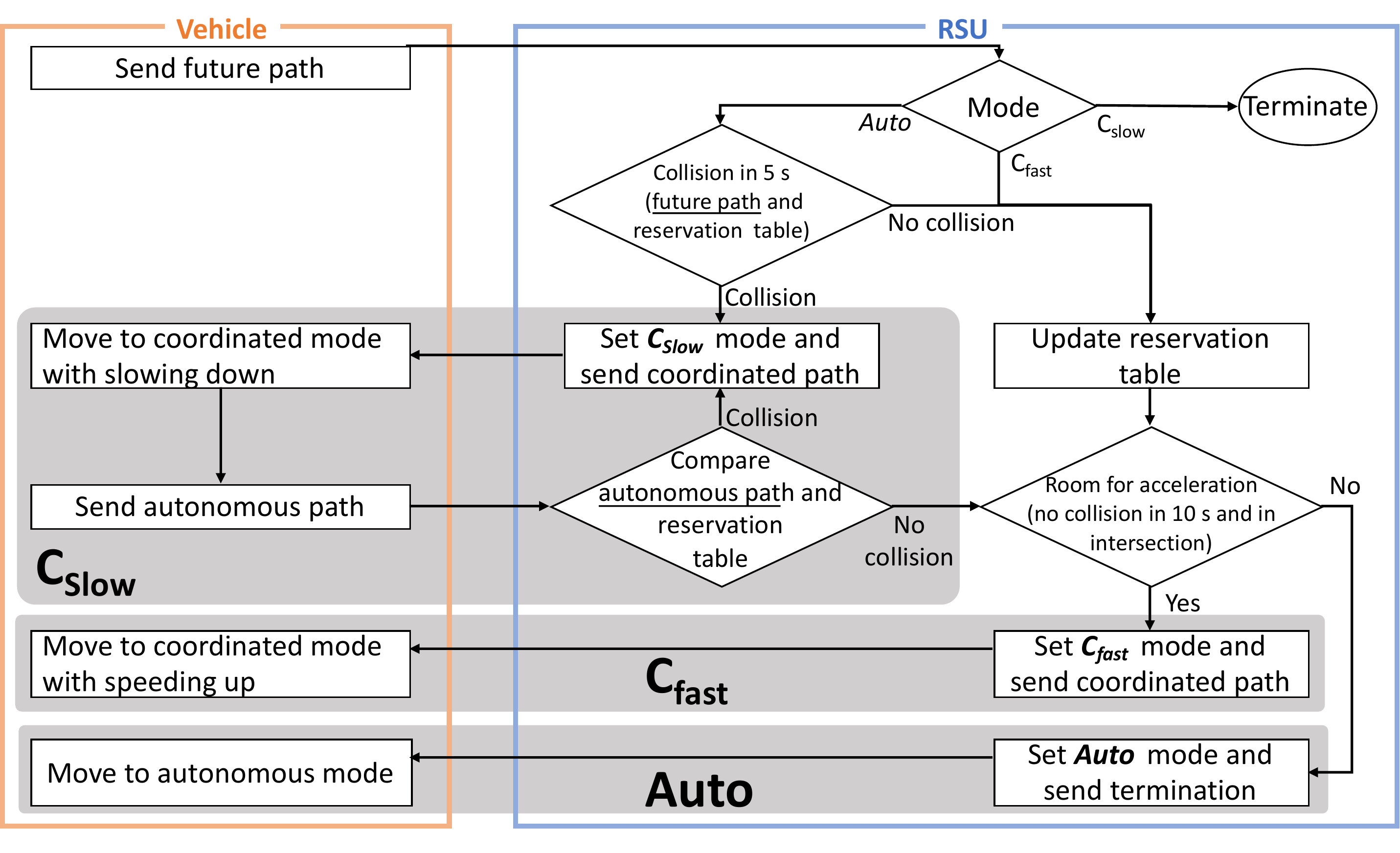}
        \caption{Mode-switching flowchart}
        \label{fig:flow}
    \end{center}
\end{figure}

The RSU applies one of three modes for each vehicle as follows: $Auto$ is used when the vehicle follows the autonomous planning mode; $C_{slow}$ is used during the coordinated mode while slowing down; and $C_{fast}$ is used during the coordinated mode while speeding up. 
The RSU chooses the modal behavior using the process shown in Fig.~\ref{fig:flow}.

During the first stage, all vehicles' modes are set to $Auto$. 
The future path of the first vehicle is then added to the table. 
% ---
When a future path is received, the RSU checks for conflicts within $T_{collision}$ seconds with all other future paths in the reservation table. 
The coordinates that exist within the safety margin $D_{margin}$ m are counted as conflicts.
In case of no conflict, the RSU then searches for an acceleration opportunity by checking if there is also no conflict in $T_{free}$ second, and the vehicle is approaching the intersection. 
In this implementation, we set $T_{collision}$ = \SI{5}{\second} to allow for enough deceleration when a conflict is detected, $T_{free}$ = \SI{10}{\second} to allow for acceleration until exiting the intersection, and $D_{margin}$ = \SI{2.8}{\meter} the width of the car.
% ---
% The RSU detects a conflict if there will be a future coordinate within \SI{2.8}{\meter} of the another vehicle’s planned path within \SI{5}{\second}. 
% In case of no conflict, the RSU then searches for an acceleration opportunity by checking if there is also no conflict in \SI{10}{\second}, and the vehicle is approaching the intersection. 
% ---

If there is room for acceleration, the RSU changes the vehicle's mode from $Auto$ to $C_{fast}$. 
To force the mode change, the RSU sends an initiation message to the vehicle to switch to $C_{fast}$. 
Then, the RSU and the vehicle generate a coordinated path that changes the speed of the autonomous vehicle to the maximum speed of the road ($V_{max}$) through the intersection as long as there is no conflict within $T_{free}$ second. 
The vehicle then follows the coordinated path and accelerates. Although the coordinated path's speed is faster than the vehicle's current speed, the control module of the Autoware accelerates smoothly until called off. 
When the vehicle passes through the intersection, the RSU changes its mode back to $Auto$ and sends a termination message to the vehicle. Thus, the acceleration and deceleration follow a smooth curve.

Alternatively, if the received future path conflicts with other paths in the reservation table, the RSU switches to $C_{slow}$. 
During this mode change, an initiation message is sent to the vehicle to switch to the $C_{slow}$ mode. After receiving the autonomous path, the RSU sends back the coordinated speed of zero. 
In the vehicle, Autoware's control module decelerates smoothly according to the current speed until the autonomous path no longer conflicts with the other paths in the reservation table (perhaps until stopping). 
During the $C_{slow}$ mode, RSU discards the vehicle's future path. 
When the vehicle path no longer conflicts with others, it calculates its room for acceleration and switches to one of the other modes (i.e., $Auto$ or $C_{fast}$).

\section{Evaluation}\label{sec:evalation}
We evaluated our implementation using the LGSVL autonomous driving simulator v.2020.06~\cite{Rong2020-ie}. 
In evaluation, we verified that the RSU intervention effectively reduced the vehicle transverse time through the intersection.

\subsection{Experimental setup and scenarios}
Three personal computers (PCs) representing \emph{Car A}, \emph{Car B}, and the \emph{RSU} were connected to the LGSVL simulator. 
The implementations included perception, planning, and controlling within the virtual world. 
We chose the Shalun map (Taiwan Car Lab Testing Facility) for a blind intersection preinstalled with LGSVL. 
A wired network connected the three PCs to the LGSVL simulator. 

\begin{figure}[tbp]
    \begin{center}
        \includegraphics[width=0.9\columnwidth]{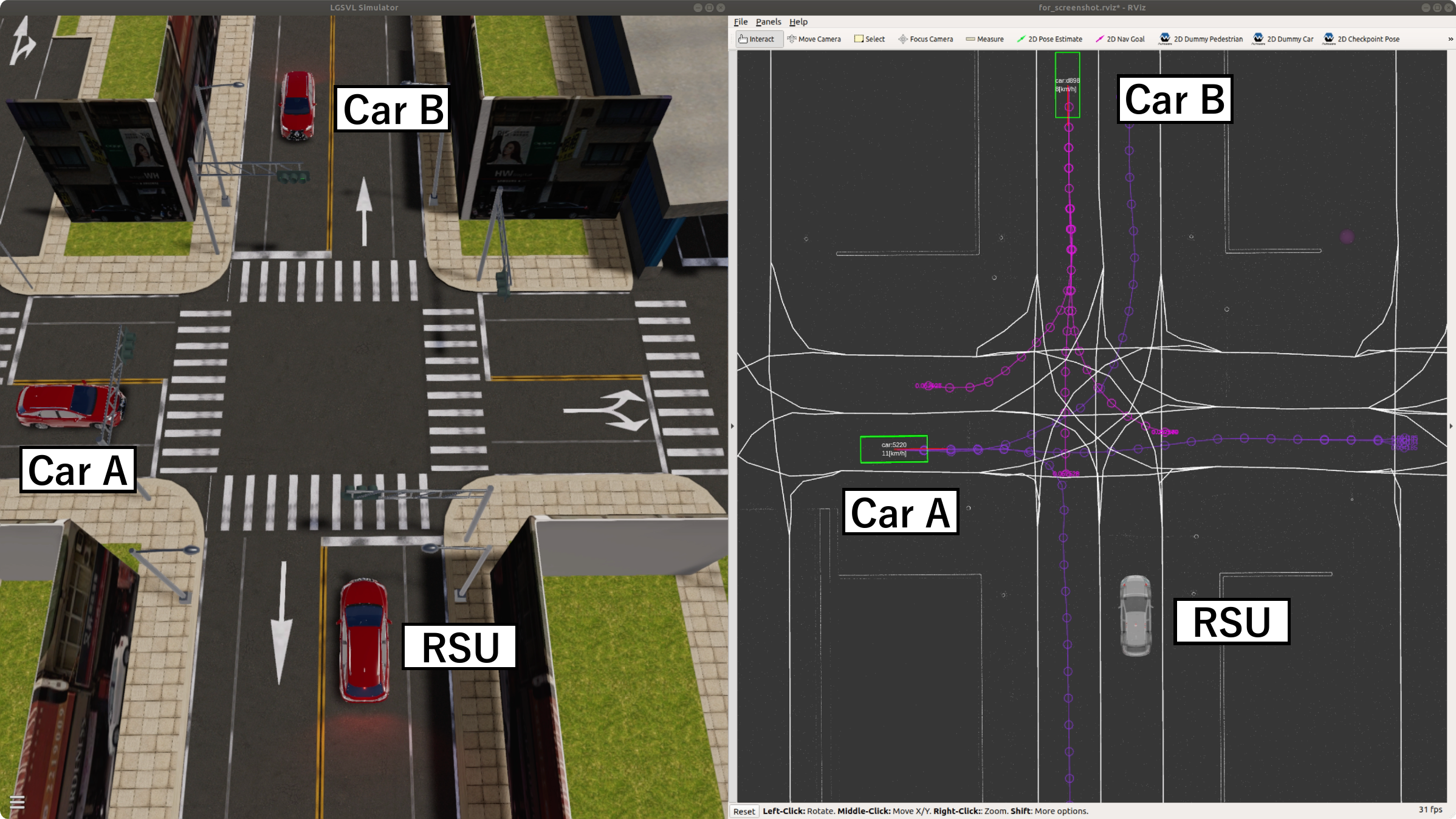}
        \caption{Screenshot of the experiment. \emph{Car A} arrives first in stand-alone mode. Left is the view of the LGSVL simulator. Right is the Rviz view.}
        \label{fig:yousu_ver3}
    \end{center}
\end{figure}

The simulation experiment was conducted at a virtual intersection having poor visibility wherein two cars (\emph{Car A} and \emph{Car B}) passed one another.
We set the starting and destination points so that \emph{Car A} would pass through the intersection before \emph{Car B} in stand-alone mode, as shown in Fig.~\ref{fig:yousu_ver3}). 
We set the maximum speed of the road $V_{max}$ = \SI{50}{\kilo\meter\per\hour}.
We set a stationary vehicle as RSU in the simulator because it had full functionality apart from the control module. 
%
% We measured the transit time from a point slightly ahead of the departure point slightly before the arrival point to not be affected by the car's acceleration when it started and the deceleration of the car when it stopped when it arrived at the destination.

We evaluated the following scenarios: 
\begin{itemize}
    \item \emph{Stand-alone}: autonomous cars without communication.
    \item \emph{Future path only}: autonomous vehicles broadcast the future path by V2V.
    \item \emph{Future path with RSU}: coordinated autonomous driving with the RSU including future path sharing. 
\end{itemize}

We conducted 10 simulation experiments for each of the three scenarios using the same starting points and destinations. 

% \begin{figure}[tbp]
%   \subfigure[View of the LGSVL simulator from RSU] {%
%       \includegraphics[width=0.61\columnwidth,clip]{figs/position7_yousu1.pdf}
%       \label{fig:position7_yousu1}
%   }%
%   \subfigure[View of Rviz from RSU] {%
%       \includegraphics[width=0.37\columnwidth,clip]{figs/position7_yousu2.pdf}
%       \label{fig:position7_yousu2}
%   }%
%   \caption{View of the experiment. \emph{Car A} arrives first in Stand-alone mode}
%   \label{fig:yousu_ver3}
% \end{figure} 

\subsection{Measurement of passing time in the intersection}
% We evaluated the proposed method using a simulator.
% As a result, it was found that by exchanging \emph{future paths}, vehicles can know the path information of other vehicles at an early stage, and each vehicle can plan its own route considering the \emph{future paths} of other vehicles.
% Besides, by using RSU to mediate each vehicle's paths, we could avoid vehicle conflicts and make traffic more efficient at intersections.

Table \ref{table:position7_origins} shows all results and passing times. 
In the stand-alone case, \emph{Car A} traversed the intersection first in all trials, as designed. 
On the other hand, \emph{Car A} traversed the first six times in the future-path-only scenario and four times in the future path with RSU. Each time, the result varied because we launched the experimental setup using a graphical user interface at slightly different timing. 

\begin{figure*}[tbp]
    \begin{center}
    \subfigure[Stand-alone] {%
        \includegraphics[width=0.3\textwidth,clip]{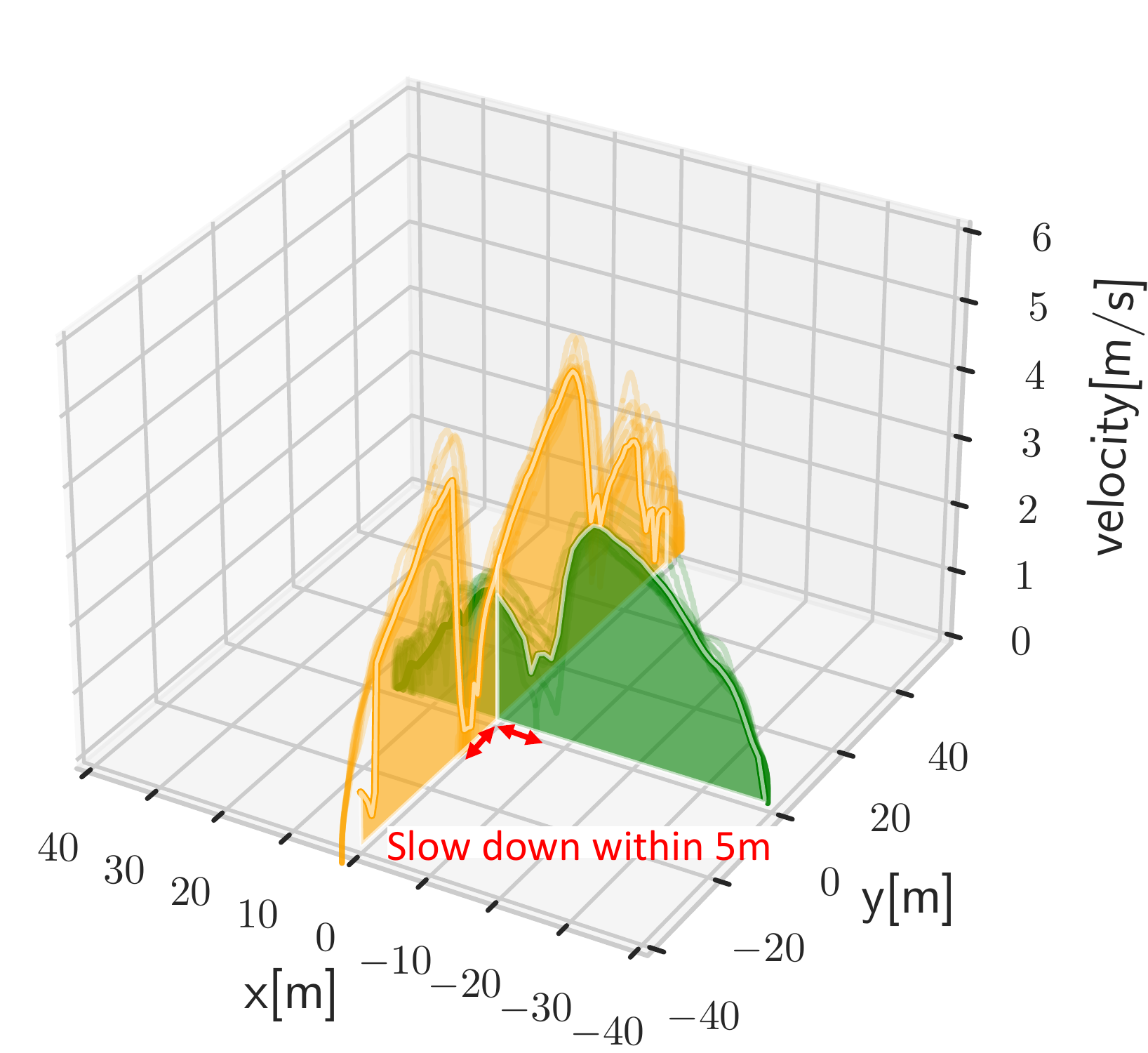}
        \label{subfig:stand-alone}
    }%
    \subfigure[Future path only] {%
        \includegraphics[width=0.3\textwidth,clip]{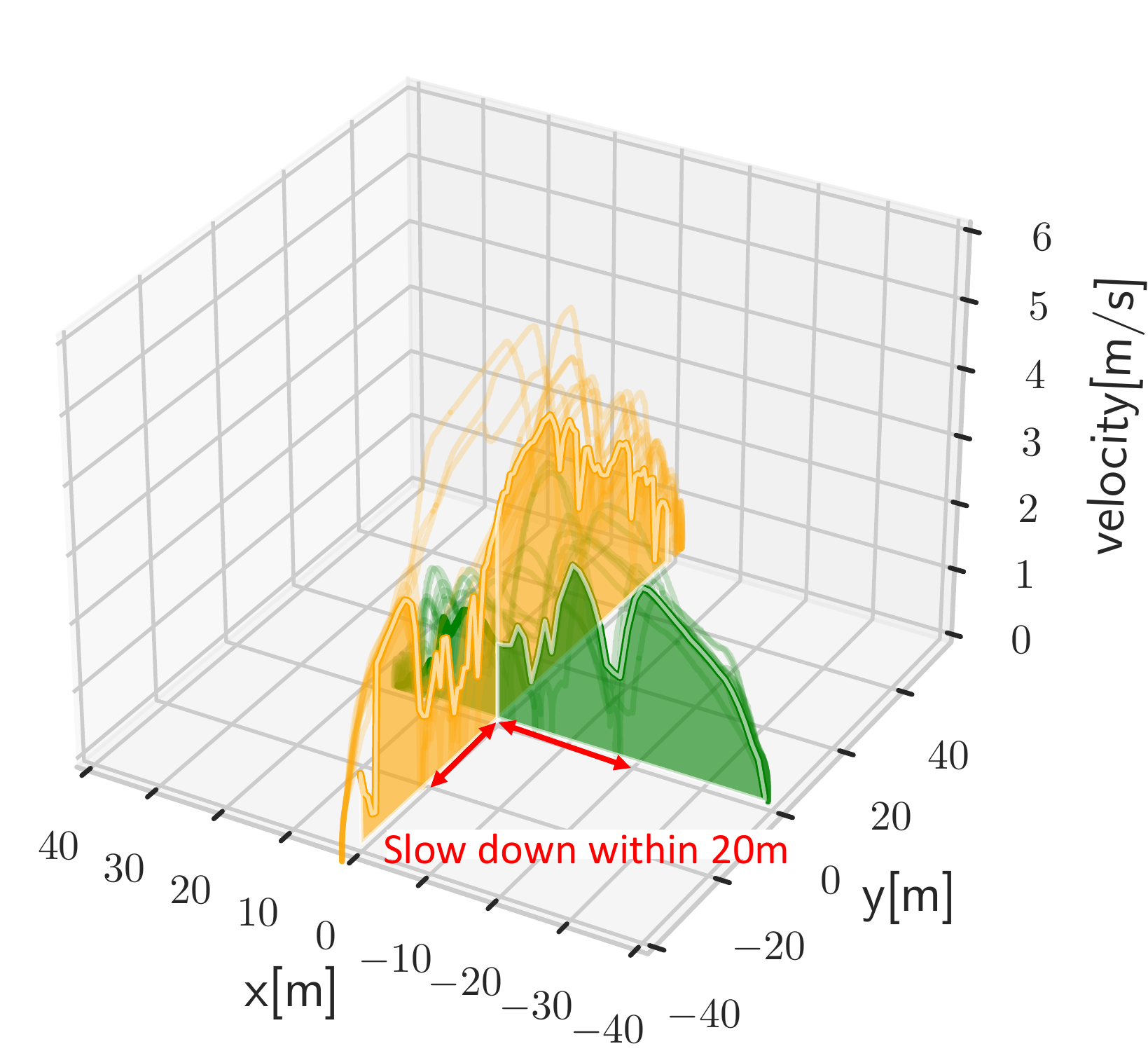}
         \label{subfig:future-path}
    }%
    \subfigure[Future path with RSU] {%
        \includegraphics[width=0.3\textwidth,clip]{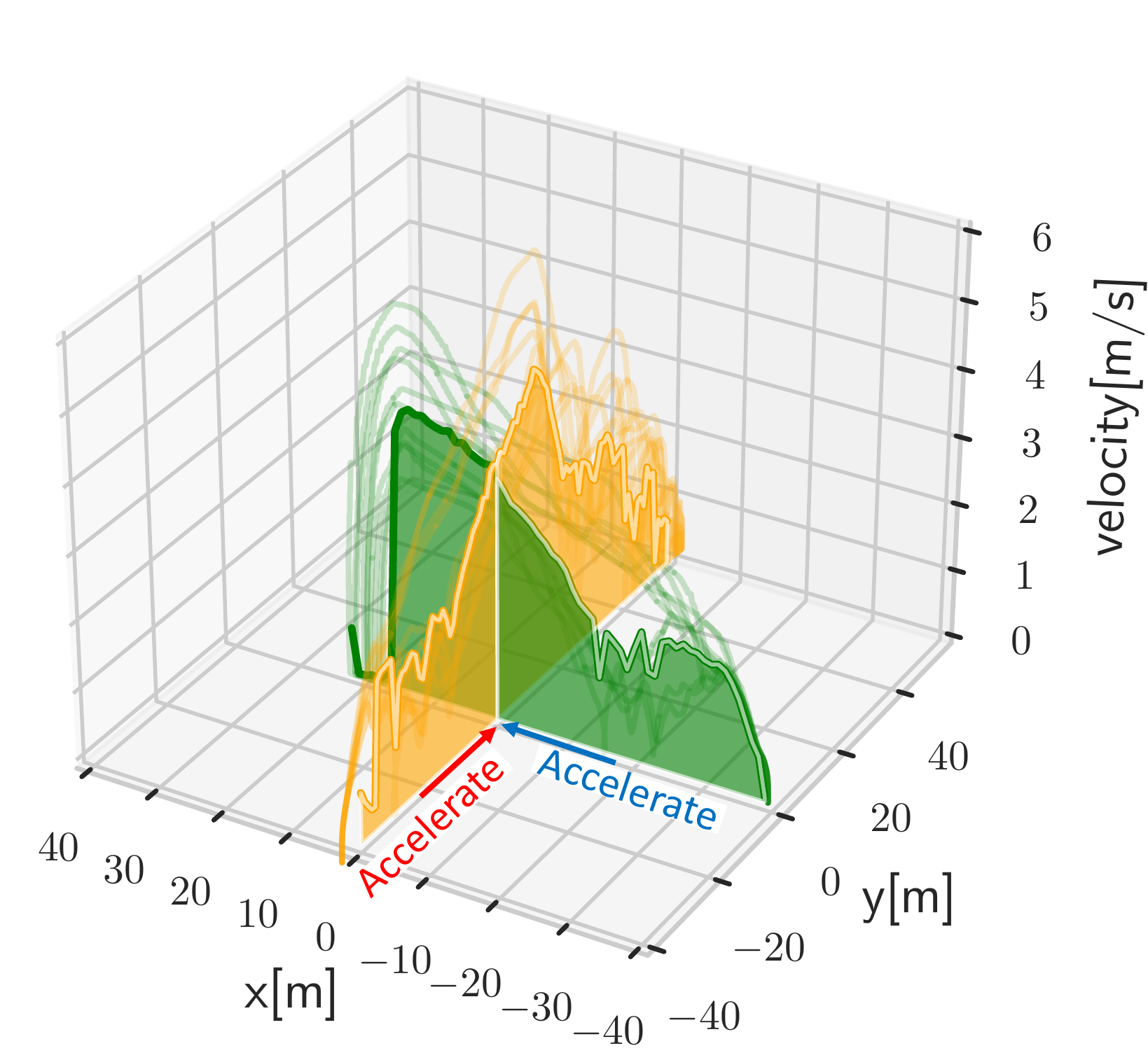}
        \label{subfig:rsu}
    }%
    \caption{Speed of two vehicles near the intersection}
    \label{fig:3d_ver7}
    \end{center}
\end{figure*}

Table~\ref{table:position7_origins} shows that the future-path-only scenario slightly increased the average passing time from \SI{36.51}{\second} to \SI{37.17}{\second}, which represents \SI{2}{\percent}. 
The reason for the worse results in the future-path-only scenario becomes apparent later by analyzing the speed of cars passing through the intersection in Figure~\ref{fig:3d_ver7}.
Our proposal (the future path with RSU) successfully reduced the average passing time from \SI{36.51}{\second} to \SI{27.05}{\second}, representing a \SI{26}{\percent} improvement. This result shows that the future paths of other vehicles were perceived as future obstacles that would otherwise degrade traffic flow. The coordinated path of the RSU effectively avoided conflict and improved flow. 

% In the absence of communication, \emph{Car A} passed through the intersection first in all cases, as shown in Table \ref{table:position7_origins}.
% At this point, according to Fig.~\ref{fig:3d_ver7}(a), \emph{Car B} brakes and stops immediately before entering the intersection. 
% \emph{Car A} also slowed down just before entering the intersection because it recognized that \emph{Car B} was coming toward it.
% As a result, it took \emph{Car A} 31.04 seconds and \emph{Car B} 41.97 seconds to pass through the intersection and reach their destinations on average, as shown in Fig.~\ref{fig:position7_origin}).
% We believe that the reason for the variation in the results was that the car was started by the GUI and the timing was different.

\begin{table}[tbp]
    \begin{center}
        \caption{Passing time in the experimental evaluation}
        \label{table:pc}
        \begin{tabular}{c|ccc} \hline
            & \multirow{2}{*}{Stand-alone} & Future  &   Future path \\ 
            &  &  Paths only &  with RSU \\ \hline \hline
            \emph{Car A} passes first & 10 & 6 & 6 \\ 
            \emph{Car B} passes first & 0 & 4 & 4 \\ \hline
            \emph{Car A} passing time & 31.04\,\si{\second} & 38.03\,\si{\second} & 23.93\,\si{\second} \\ 
            \emph{Car B} passing time & 41.97\,\si{\second} & 36.32\,\si{\second} & 30.18\,\si{\second} \\ 
            Average passing time& 36.51\,\si{\second} & 37.17\,\si{\second} & 27.05\,\si{\second} \\ \hline
        \end{tabular}
        \label{table:position7_origins}
    \end{center}
\end{table}

\begin{figure}[tbp]
    \begin{center}
        \includegraphics[width=0.8\columnwidth]{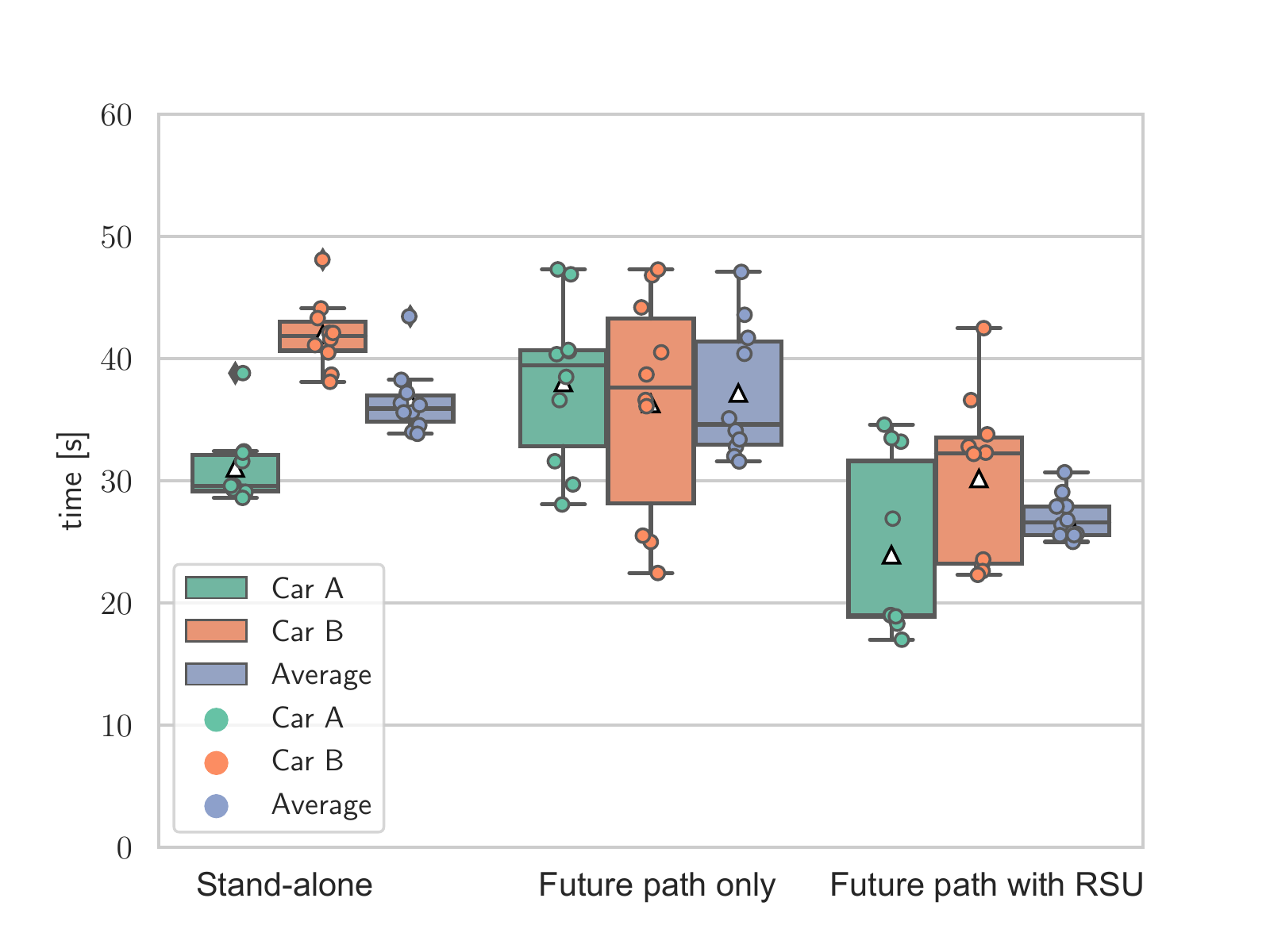}
        \caption{Passing times of 10 results}
        \label{fig:position7_origin}
    \end{center}
\end{figure}

Figure~\ref{fig:position7_origin} shows the passing times of the 10 trials and their variations. 
In the stand-alone case, the results showed only a few variations because \emph{Car A} always passed first. 
When the vehicles exchanged future paths, the passing times varied because \emph{Car A} traversed the first six times and \emph{Car B} four. 
As a result, \emph{Car A} took \SI{6.99}{\second} longer to traverse than in the stand-alone case (\SI{22}{\percent} slower), whereas \emph{Car B} arrived \SI{5.65}{\second} earlier (\SI{13}{\percent} faster). 
Our proposed method improved the passing times \SI{23.0}{\percent} and \SI{28.1}{\percent} for \emph{Car A} and \emph{Car B}, respectively. The proposed method reduces traversal times by \SI{25.9}{\percent} with a few variations. 

% According to Fig.~\ref{fig:3d_ver7}(b), \emph{Car A}, and \emph{Car B} recognized the other car and started to slow down earlier than in the case without communication.
% As shown in Fig.~\ref{fig:position7_origin}, it can pass through the intersection and reach the destination. 
% As a result, as shown in Fig.~\ref{fig:position7_origin}, it took \emph{Car A} 38.03 seconds and \emph{Car B} 36.32 seconds to pass through the intersection and reach the destination. 
% \emph{Car A} took 6.99 seconds more time to pass through than when there was no communication.
% On the other hand, \emph{Car B} took 5.65 seconds longer than the case without communication. \emph{Car B}, on the other hand, was able to pass through 5.65 seconds faster than without communication.
% When the RSU intervened, \emph{Car A} moved ahead six times and \emph{Car B} four times, as shown in Table \ref{table:position7_origins}.
% According to Fig.~\ref{fig:3d_ver7}(c), when the RSU intervened, the car that sent the route through the intersection to the RSU first accelerated, while the car that sent it later started to decelerate earlier.
% As a result, as shown in Fig.~\ref{fig:position7_origin}, it took \emph{Car A} and \emph{Car B} to pass through the intersection and reach their destinations with 23.93 seconds and 30.18 seconds, respectively, and \emph{Car A} was able to pass through the intersection 7.11 seconds faster than without communication. \emph{Car B} took 30.18 seconds.
% \emph{Car B} was able to pass through 11.79 seconds faster than when there was no communication.

Figure~\ref{fig:3d_ver7} shows the speed of the two vehicles near the intersection. 
Green is the result for \emph{Car A} and orange for \emph{Car B}, and the larger value indicates the direction of travel. 
The transparent colored line shows the speed of each trial, and the solid line shows the average speed. 
$(x,y)=(0,0)$ represents the intersection.

As depicted in Fig.~\ref{subfig:stand-alone}, in the stand-alone scenario, \emph{Car B} always stopped at \SI{5}{\meter} before the intersection for \emph{Car A} passing first. 
As shown in Fig.~\ref{subfig:future-path}, in the future path-only scenario, \emph{Car A} and \emph{Car B} recognized each other. They slowed sooner (within \SI{20}{\meter} from the intersection) than in the stand-alone case, which explains why the future path exchange degrades the passing time. 
Figure~\ref{subfig:rsu} shows that both \emph{Car A} and \emph{Car B} did not stop before the intersection using our proposal. 
The first vehicle that sent its future path to the RSU accelerated along the coordinated path, and the other vehicle decelerated sooner than in the stand-alone case. 

% According to Fig.~\ref{fig:3d_ver7}(c), when the RSU intervened, the car that sent the route through the intersection to the RSU first accelerated, while the car that sent it later started to decelerate earlier.
% As a result, as shown in Fig.~\ref{fig:position7_origin}, it took \emph{Car A} and \emph{Car B} to pass through the intersection and reach their destinations with 23.93 seconds and 30.18 seconds, respectively, and \emph{Car A} was able to pass through the intersection 7.11 seconds faster than without communication. \emph{Car B} took 30.18 seconds.
% \emph{Car B} was able to pass through 11.79 seconds faster than when there was no communication.

% 10:45:45 From Xavier Defago to Everyone : https://stackoverflow.com/questions/20407936/matplotlib-not-displaying-intersection-of-3d-planes-correctly
% 10:48:02 From Manabu Tsukada to Everyone : https://stackoverflow.com/questions/28531752/display-the-maximum-surface-in-matplotlib/28608098#28608098
% 10:50:38 From Xavier Defago to Everyone : tikz
% 10:50:54 From Xavier Defago to Everyone : https://en.wikipedia.org/wiki/PGF/TikZ
% 10:51:37 From Xavier Defago to Everyone : https://en.wikipedia.org/wiki/OmniGraffle  

\subsection{Discussion}\label{subsec:discussion}

Unlike this experiment, actual cars would communicate via wireless networks. 
Therefore, the network resources are limited, and we need to consider the network load.
First of all, vehicles and RSUs can send a maximum of 120 trajectory points, theoretically limited by the maximum UDP packet size (1460 Bytes), assuming that the latitude, longitude, and time are 32bits~\cite{ETSI-EN-302-636-4-1-GN}. 
It consumes the network bandwidth equivalent to 116 Kbps when the maximum UDP packet is sent at 10 Hz. 
In the proposed method, future paths, autonomous paths, and coordinated paths are sent at 10 Hz, which requires 350 Kbps of network traffic per car. 
The proposed method can control up to 17 cars when the data rate of wireless media is 6.0 Mbps (i.e., DSRC). 
Some optimizations are necessary to coordinate more cars, such as reducing the number of trajectory points, transmission rate, etc. 
According to~\cite{Da_CML2017-rm}, fully automated, where all maneuvers are performed cooperatively, is classified as Day 4 (to be deployed by 2040-2045). 
Thus, the wireless media available at that time may solve the problem.

In addition, our proposal is a method for RSUs to take over the control of automobiles. It may be challenging to obtain the consent of current manufacturers of autonomous self-driving cars. 
Therefore, our method may be adapted in the situation where the same manager controls both cars and RSUs, such as factories.

\section{Conclusion and future work}\label{sec:conclusion}

In cooperative ITS, autonomous vehicles use V2X to exchange real-time information.
However, real-time path planning becomes inefficient when driving paths conflict.
This study proposed a method for cooperative path planning using future path sharing that considers future planning information.
We placed an RSU at the intersection to coordinate path planning among two autonomous vehicles and implemented our proposition by extending Autoware, an open-source autonomous driving software, to simulate the traversal times of two vehicles at a blind intersection using the LGSVL autonomous driving simulator. 
The results show that future-path broadcasts improve traversal times because the autonomous vehicles adjust their speeds based on their combined future paths, governed by the RSU. Our proposal effectively coordinates vehicle speeds and achieves a \SI{26}{\percent} faster traversal time for both vehicles through the intersection. 

Future work will include a large-scale simulation involving many cars and pedestrians. We used a wired network in this study. However, we plan to use a wireless network or a network simulator in the future. Furthermore, field experiments would provide a more realistic environment for evaluation. Furthermore, it is necessary to manage vehicles' priority at intersections and compare the performance with related works.
The proposed scheme can be extended to mixed traffic stream non-connected cars and vulnerable road users (e.g., cyclists and pedestrians).  RSU's prediction module can add the predicted paths of the dynamic objects to the reservation table and generate the conflict-free coordinated path for autonomous vehicles. The evaluation for the scenario is future work.

% TODO: 議論
% - 曲がる車がない
% - ネットワークに関する評価

% Future workに優先順位をつける。もしくは、重要なところだけ書く
% やると面白いところを書く。

% PageReduction
\section*{Acknowledgment}

This work was partly supported by JSPS KAKENHI (grant number: 19KK0281 and 21H03423).

\bibliographystyle{IEEEtran}
\bibliography{main}

\end{document}